\documentclass[10pt,twocolumn,letterpaper]{article}

\usepackage[pagenumbers]{cvpr} 

\usepackage{graphicx}
\usepackage{amsmath}
\usepackage{amssymb}
\usepackage{booktabs}
\usepackage{comment}

\usepackage[pagebackref,breaklinks,colorlinks]{hyperref}

\newcommand{\OURS}{ObjectMatch}

\usepackage[capitalize]{cleveref}
\crefname{section}{Sec.}{Secs.}
\Crefname{section}{Section}{Sections}
\Crefname{table}{Table}{Tables}
\crefname{table}{Tab.}{Tabs.}

\begin{document}


\title{\OURS{}: Robust Registration using Canonical Object Correspondences}
\vspace{-1cm}
\author{Can G{\"u}meli 
\qquad\qquad
Angela Dai
\qquad\qquad
Matthias Nie{\ss}ner \\
\\
Technical University of Munich 
}

\begin{figure}
\twocolumn[{
\renewcommand\twocolumn[1][]{#1}%
\maketitle
\begin{center}
    \centering
    \includegraphics[width=.98\textwidth]{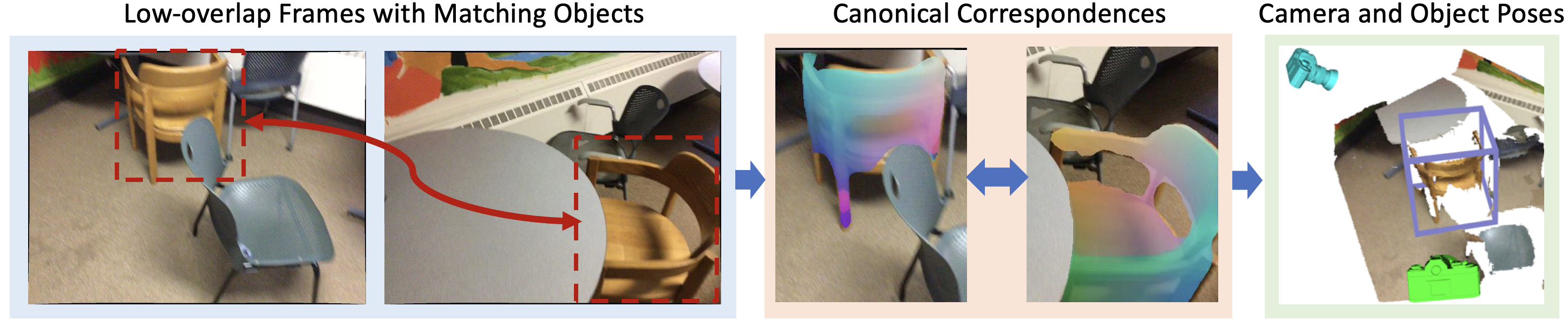}
    \vspace{-2mm}
    \caption{
    Modern camera pose estimation relies on feature matching between overlapping frames - in this work, we present \OURS{} to find correspondences between frames with little or no overlap by predicting semantic mappings through canonical object correspondences. The images above share no direct overlap, yet our method establishes indirect correspondences, thus enabling a successful registration.
    }
    \label{figure:teaser}

\end{center}
}]
\end{figure}

\begin{abstract}
\vspace{-2mm}
We present \OURS{}\footnote{ \url{https://cangumeli.github.io/ObjectMatch/}}, a semantic and object-centric camera pose estimator for RGB-D SLAM pipelines. 
Modern camera pose estimators rely on direct correspondences of overlapping regions between frames; however, they cannot align camera frames with little or no overlap.
In this work, we propose to leverage indirect correspondences obtained via semantic object identification.
For instance, when an object is seen from the front in one frame and from the back in another frame, we can provide additional pose constraints through canonical object correspondences.
We first propose a neural network to predict such correspondences on a per-pixel level, which we then combine in our energy formulation with state-of-the-art keypoint matching solved with a joint Gauss-Newton optimization.
In a pairwise setting, our method improves registration recall of  state-of-the-art feature matching, including from 24\% to 45\% in pairs with 10\% or less inter-frame overlap. 
In registering RGB-D sequences, our method outperforms cutting-edge SLAM baselines in challenging, low-frame-rate scenarios, achieving more than 35\% reduction in trajectory error in multiple scenes.
\end{abstract}

\section{Introduction}
\label{sec:intro}
RGB-D registration and 3D SLAM has been a fundamental task in computer vision, with significant study and enabling many applications in mixed reality, robotics, and content creation. 
Central to both state-of-the-art traditional and learning-based camera pose estimation is establishing correspondences between points in input frames.
However, correspondence estimation remains quite challenging when there is little or no overlap between frames.

In contrast, humans can easily localize across these challenging scenarios by leveraging additional semantic knowledge -- in particular, by further localizing at the level of objects and identifying matching objects between views.
For instance, when observing a chair from the back and the side (e.g., in Figure~\ref{figure:teaser}), view overlap is minimal (or even no view overlap), resulting in failed registration from keypoint matching. 
However, the semantic knowledge of the chair and its object pose nonetheless enables humans to estimate the poses from which the front and side views were taken.
Thus, we propose to take a new perspective on camera pose estimation and imbue camera registration with awareness of such semantic correspondences between objects for robust performance in these challenging scenarios.

To this end, we propose \OURS{}, a new paradigm for camera pose estimation leveraging canonical object correspondences in tandem with local keypoint correspondences between views.
This enables significantly more robust registration under a variety of challenging scenarios, including low view overlap.
For a sequence of input frames, \OURS{} learns to semantically identify objects across frames, enabling a compact, global parameterization of 9-DoF object poses.
Object correspondences are established through predicting normalized object coordinates~\cite{noc}, dense correspondences from object pixels to a canonically oriented space for each object.
We then formulate a joint camera and object pose optimization that constrains object correspondences indirectly, operating irrespective of the shared visibility of image regions. 
Our approach is complementary to state-of-the-art SLAM methods, and we leverage our energy formulation to complement state-of-the-art keypoint matching~\cite{superpoint, superglue, bundlefusion} in a joint Gauss-Newton optimization.

Our method outperforms strong baselines in both pairwise registration and registration of RGB-D frame sequences. 
In pairwise registration of challenging ScanNet~\cite{scannet} image pairs, we improve pose recall from 24\% to 45\% when the overlap is below 10\%.
On sequence registration of room-scale RGB-D scenes, our method outperforms various strong baselines in difficult, low-frame-rate settings in several TUM-RGBD~\cite{tum_rgbd} and ScanNet~\cite{scannet} scenes, reducing the trajectory error by more than 35\% in multiple challenging scenes.

To sum up, our main contributions include:

\begin{itemize}
    \vspace{-3mm}\item An object-centric camera pose estimator that can handle low-overlap frame sets via indirect, canonical object correspondences established with predicted dense, per-pixel normalized object coordinates. 
    \vspace{-2mm}\item A joint energy formulation that leverages semantic object identification and dense, normalized object coordinates corresponding to canonical object geometries.
    \vspace{-6mm}\item Our semantic grounding of object correspondences enables significantly more robust registration in low-overlap and low-frame-rate cases. \OURS{} improves over state of the art from 24\% to 45\% registration recall of $\leq 10\%$ overlap frame pairs and achieves over 35\% trajectory error reduction in several challenging sequences.
\end{itemize}

\section{Related Work}
\noindent\textbf{RGB-D Registration and SLAM.} In recent years, there have been many advances in indoor RGB-D reconstruction. Earlier RGB-D fusion approaches focus on frame-to-model camera tracking~\cite{kinectfusion, voxelhashing}. To handle the loop closures better, more recent SLAM systems introduce explicit strategies or global optimization methods for handling loop closures through global optimization~\cite{global_reg, bundlefusion, elasticfusion, bad_slam, orb_slam2} to fix tracking errors. More recently, deep learning techniques have been applied to registration and SLAM scenarios, with methods ranging from geometric point cloud registration~\cite{deep_global_reg, predator, geotrans} as well as neural field based SLAM techniques 
\cite{imap, di_fusion, nice_slam}. Despite all the successes in RGB-D SLAM and registration, the task is still challenging since incomplete loop closures observed via low-overlap frames cannot be handled, and most SLAM methods require a very high overlap between consecutive frames to track cameras accurately.

\medskip\noindent\textbf{Feature Matching.} Modern RGB(-D) camera pose estimators rely on a feature-matching backbone. Classical global registration techniques~\cite{global_reg, fast_global_reg} use FPFH features~\cite{fpfh} over point cloud fragments. On the other hand, many global RGB-D SLAM techniques rely on sparse color features~\cite{bundlefusion, orb_slam2}. While being successful in many scenarios, conventional feature matching often fails when the inter-frame overlap is low. Therefore, deep learning techniques have been utilized for predicting overlapping regions based on geometry or color. On the geometric side, Deep Global Registration~\cite{deep_global_reg} predicts overlapping point features using nearest neighbor search over learned geometric features~\cite{fcgf}. Methods such as PREDATOR~\cite{predator} and  Geometric Transformer~\cite{geotrans} use attention mechanisms to target overlapping regions for registration. In the domain of color features, SuperPoint and SuperGlue~\cite{superpoint, superglue} build a formative approach in GNN-based keypoint feature matching. Methods such as LoFTR~\cite{loftr} introduce more dense and accurate sub-pixel level matching. Despite being very successful in handling wide-baseline scenarios, learned feature matching still requires a significant amount of shared visibility and geometric overlap.

\medskip\noindent\textbf{Camera Pose Estimation with Semantic Cues.} Several methods have been developed to incorporate semantic priors to improve low-overlap registration. PlaneMatch~\cite{planematch} proposed coplanarity priors for handling loop closures and low-overlap cases. Our method instead leverages object-centric constraints, exploiting the power of semantic object recognition. 
Another related direction is feature hallucination by leveraging image and object semantics.
NeurHal \cite{neurhal} focuses on correspondence hallucination using image inpainting and outpainting, formulating a PnP optimization over hallucinated matches.
Virtual Correspondence (VC)~\cite{virtual_cor} introduces a human-centric approach that leverages hallucinated object volumes to form virtual epipolar constraints between frames.
In contrast, we use indirect instead of direct correspondences that do not require hallucinated object volumes or image regions. Furthermore, our method works on a diverse set of furniture categories while VC focuses on humans. Pioneered by SLAM++~\cite{slam_pp}, there is also a rich literature of object-centric SLAM solutions, e.g.,~\cite{node_slam, wide_disparity}. Such SLAM methods leverage local, per-frame poses of objects to establish constraints; instead, we develop a global object pose optimization that is more robust against occluded viewpoints.

\medskip\noindent\textbf{Object Pose Estimation using Normalized Object Coordinates.} 3D object pose estimation has been widely studied from RGB and RGB-D inputs.
Normalized Object Coordinate Space (NOCS)~\cite{noc} was proposed to form dense correspondences from input RGB-D frames to canonical object geometries, enabling  better generalization than direct regression of object poses.
End2End SOCs~\cite{end2end} formulated a NOC-based approach for CAD retrieval and alignment to 3D scans, using a differentiable Procrustes algorithm. 
To enable CAD alignment to single RGB images, ROCA~\cite{roca} leveraged NOC space in combination with predicted depths, formulating a robust differentiable Procrustes optimization~\cite{roca}. Seeing Behind Objects~\cite{seeing_behind_objects} further leveraged NOC correspondences both to obtain local object poses and object completion for re-identification for RGB-D multi-object tracking. Wide Disparity Re-localization ~\cite{wide_disparity} uses the NOC predictions from~\cite{noc} to construct an object-level map for re-localization in SLAM. In contrast to these approaches that focus on individual object poses, we use NOC correspondences directly in a multi-frame, global camera, and object pose optimization.

\section{Method}

\begin{figure}
\twocolumn[{
\renewcommand\twocolumn[1][]{#1}%
\begin{center}
    \centering
    \includegraphics[width=.97\textwidth]{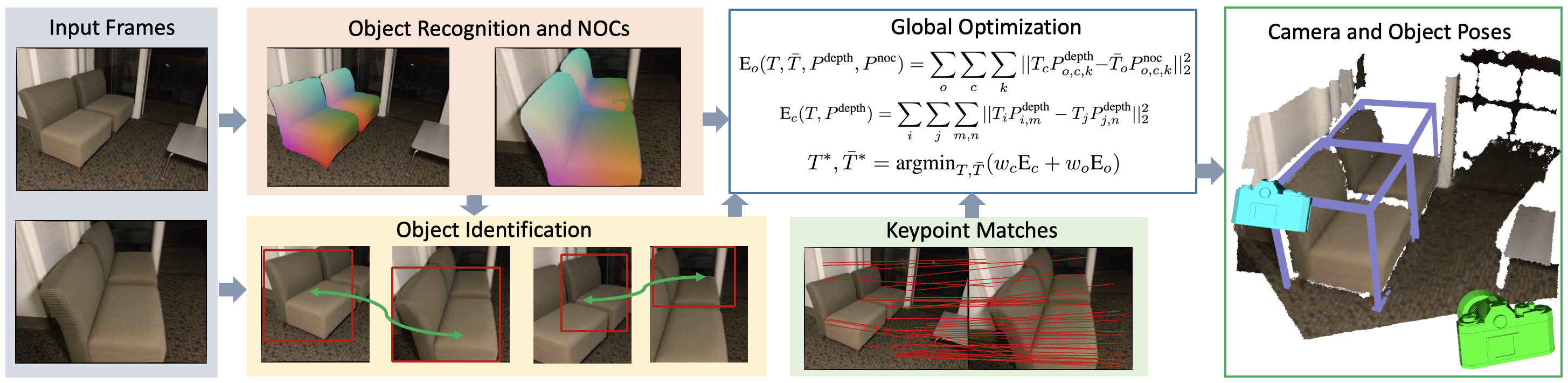}
    \vspace{-2mm}
    \caption{Overview of our approach to incorporate object correspondence grounding in global pose estimation. From a set of input RGB-D frames, \OURS{} predicts object instances for each frame with dense normalized object correspondences. The predicted object instances are used to identify objects across frames, forming indirect object correspondences. We combine object correspondences with SuperGlue \cite{superpoint, superglue} keypoint matches in a joint energy optimization that yields both camera and object poses in a global registration.
    }
    \label{fig:pipeline}
\end{center}
}]
\end{figure}

\subsection{Problem Setup}
Given $K$ RGB-D frames $\{(I_1^c, I_1^d), ..., (I_K^c, I_K^d)\}$, we aim to optimize their 6-DoF camera poses $T_{c}{ }={ }\{T_2, ..., T_K\}$, assuming the first frame is the reference, i.e., $T_1=\mathbb{I}$.
A 6-DoF camera pose $T_i$ is represented by Euler angles $\gamma$ and translations $t$, $T_i = (\gamma_x, \gamma_y, \gamma_z, t_x, t_y, t_z)$. 

We also parameterize global, 9-DoF object poses, $\bar{T}_o = (\gamma_{x}, \gamma_{y}, \gamma_{z}, t_{x}, t_{y}, t_{z}, s_{x}, s_{y}, s_{z})$, comprising 6-DoF angles and translations, and 3-DoF anisotropic scales $s$.

We formulate a joint energy optimization of the form:
\begin{equation} \label{optim_high_level}
    T^{*}, \bar{T}^{*} = \textrm{argmin}_{T, \bar{T}}(\textrm{E}_c(T, M) + \textrm{E}_o(T, \bar{T}, N))
\end{equation}
where $M$ are inter-frame feature matches and $N$ are intra-frame canonical object correspondences established with normalized object coordinates (NOCs) that densely map to the canonical space of an object in $[-0.5, 0.5]^3$.
$\textrm{E}_c$ is the feature-matching energy function, and $\textrm{E}_o$ is our object-centric energy function. Since robust feature matching and optimization are readily available off the shelf~\cite{superpoint, superglue, bundlefusion, geotrans}, our method focuses on building the object-centric energy $\textrm{E}_o$.{ }To this end,  we need two function approximators, realized via deep neural networks:
(1) a learned model for object recognition and NOC prediction, and 
(2) a learned model for object identification.
The realization of these networks is described in Sections~\ref{method:noc} and \ref{method:match}, respectively, and energy function Eq.~\ref{optim_high_level} in Section~\ref{method:opt}.
An overview of our approach is visualized in Figure~\ref{fig:pipeline}.

\subsection{Predicting Object Correspondences} \label{method:noc}
\begin{figure}
    \centering
    \includegraphics[width=0.45\textwidth]{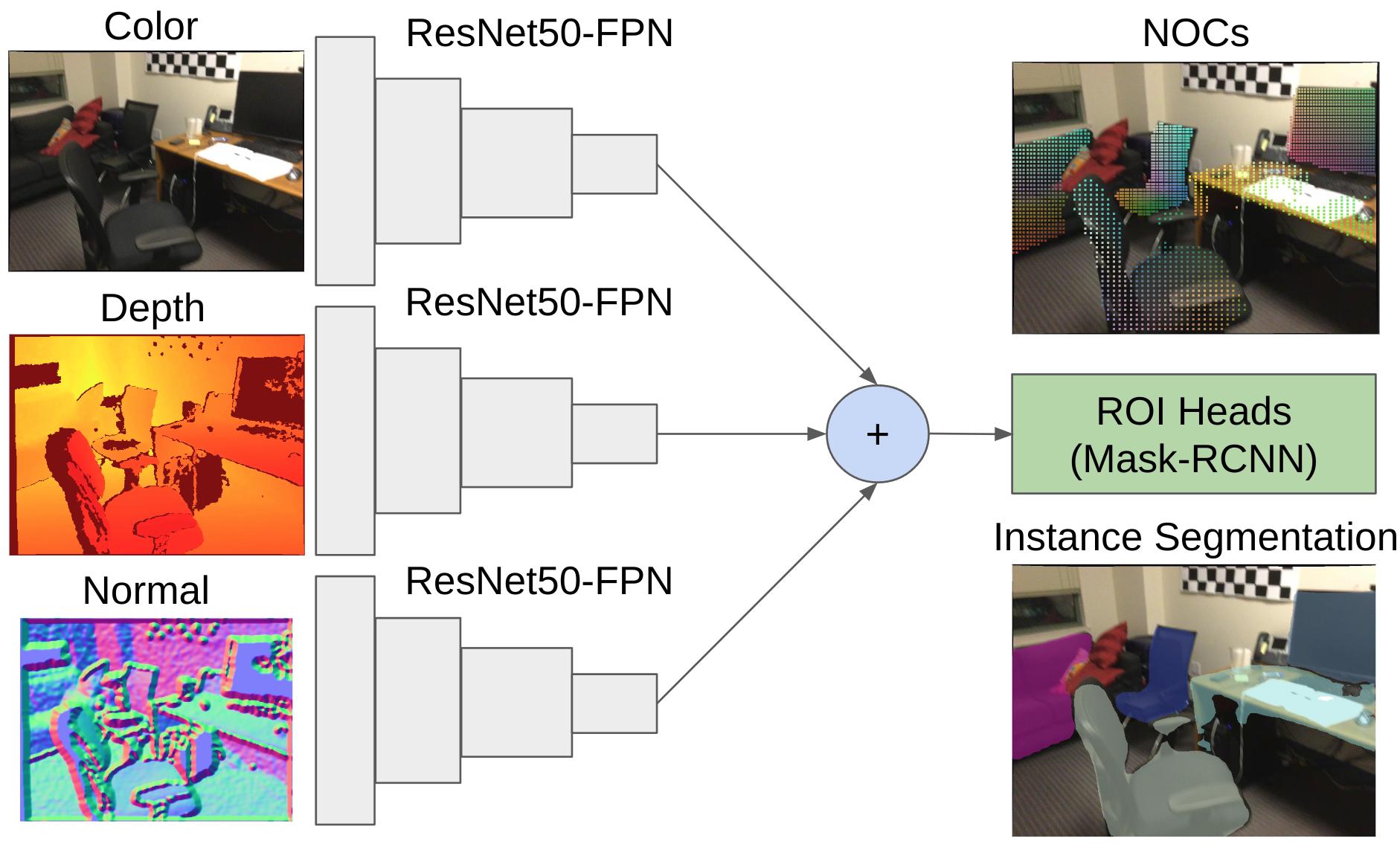}
    \vspace{-2mm}
    \caption{Multi-modal object recognition and NOC prediction. Our ResNet-FPN \cite{resnet, fpn} backbone takes color, reversed jet colored depth \cite{jet_multimodal}, and 2x downsampled colored 3D depth normals, and produces multi-scale features by averaging different input encodings. From the obtained features, our method recognizes objects and predicts NOCs for each object, based on a Mask-RCNN \cite{maskrcnn}-style prediction. }
    \label{fig:noc_pred}
\end{figure}

To obtain object constraints in our final optimization, we recognize objects via object detection and instance segmentation, and predict object correspondences as dense NOCs~\cite{noc} for each object, as shown in Figure~\ref{fig:noc_pred}.
We build on a Mask-RCNN~\cite{maskrcnn} with ResNet50-FPN~\cite{resnet, fpn} backbone, pre-trained on ImageNet~\cite{imagenet} and COCO~\cite{coco}.

To input the depth of the RGB-D frames, we propose a modified, multi-modal ResNet50-FPN \cite{resnet, fpn, jet_multimodal} backbone. Our backbone takes 480x640 color, 480x640 reverse jet-colored depth, and 240x320 colored depth normals as input. 
We average the resulting FPN features to a single feature pyramid:
\begin{equation}
    G= \frac{\mathrm{FPN}^{c}(I^c) + \mathrm{FPN}^d(I^{d}) + \mathrm{U}(\mathrm{FPN^n}(I^{n}))}{3},
\end{equation}
where $I^c, I^d, I^n$ are color, depth, and normal images, $\mathrm{FPN}^c, \mathrm{FPN}^d, \mathrm{FPN}^n $ are the corresponding ResNet50-FPN backbones, and $\mathrm{U}$ is an upsampling operator to match the normal features' spatial size with others. 
This enables fine-tuning the pre-trained bounding box, class, and instance segmentation heads of Mask-RCNN \cite{maskrcnn}, while also exploiting  depth information. 
We use symmetrically structured FPNs, all pre-trained on ImageNet and COCO as initialization, but without any parameter sharing.

To obtain object correspondences, we establish mappings from detected objects to their canonical spaces, in the form of dense NOCs.
That is, for each pixel in the object's instance mask, we predict 3D points $P^{\textrm{noc}}_o$ using a fully convolutional network:
\begin{equation}
    P^{\textrm{noc}}_o = \textrm{FCN}(G_o),\;\; p\in [-0.5, 0.5]^3\; \forall\; p\in P^{\textrm{noc}}_o.
\end{equation}
We optimize an $\ell_1$ loss $L_{\textrm{noc}}$ using ground-truth NOCs $P^{\textrm{noc-gt}}$,
\begin{equation}
    L_{\textrm{noc}} = \sum_o\sum_i||P^{\textrm{noc}}_{o,i} -P^{\textrm{noc-gt}}_o||_1.
\end{equation}

Since symmetric objects~\cite{scan2cad, noc} induce ambiguities in NOCs (e.g., a round table), we classify symmetry type of objects (round, square, rectangle, non-symmetric), $c_{\textrm{sym}} = \textrm{MLP}^{\textrm{sym}}(G_o)$, optimized using a cross-entropy loss $L_{\textrm{sym}}$. We also make $L_{\textrm{noc}}$ symmetry aware, taking the minimum $\ell_1$ difference over the set of correct NOCs~\cite{noc}. We use non-symmetric objects during inference to avoid inconsistent NOCs across views.

In addition to NOCs, we also regress anisotropic 3D object scales $s_{o}$ using a fully connected network, $s_o = \textrm{MLP}^{\textrm{scale}}(G_o)$, and optimize $s_o$ with an $\ell_1$ loss $L_{\textrm{scale}}$.
The object scale enables holistic object pose understanding within each frame and helps to filter potential object matches across views using scale consistency.

Finally, to make our NOC-depth correspondences least-squares friendly for our desired optimization, we also introduce a per-frame differentiable Procrustes objective $L_{\textrm{proc}}$, using a differentiable Kabsch solver \cite{pytorch3d} to obtain local object rotations and translations:
\begin{equation}
R_o^{*}, t_o^{*} = \textrm{argmin}_{R_o, t_o}(\sum_i||R_o (P^{\textrm{noc}}_{o,i} \odot s_o) + t_o - P^{\textrm{depth}}_{o,i}||_2^2)
\end{equation}
for each object $o$, where  $P^{\textrm{depth}}_{o}$ are back-projected input RGB-D depths corresponding to the object's predicted instance mask in its region of interest, and $\odot$ denotes element-wise multiplication. We train the local object poses with
\begin{equation}
L_{\textrm{proc}} = w_r\sum_o||R_o^{*} - R_o^{\textrm{gt}}||_1 + w_t\sum_o||t_o^{*} - t_o^{\textrm{gt}}||_2^2.
\end{equation}

Our full loss used for training is then
\begin{equation}
    L = L_{m} + w_{n}L_{\textrm{noc}} + w_{s}L_{\textrm{scale}} + w_{\textrm{sym}}L_{\textrm{sym}} + w_{p}L_{\textrm{proc}},
\end{equation}
where $L_m$ is the sum of Mask-RCNN losses \cite{maskrcnn} and $w$ are scalar weights balancing additional losses.

\smallskip
\noindent \textbf{Implementation.}
We use the augmented 400k ScanNet train image split for training \cite{scannet, roca}, with  Scan2CAD labels of the most common 9 furniture categories following \cite{scan2cad, vid2cad, roca}.
We train a standard Detectron2 Mask-RCNN pipeline \cite{maskrcnn, detectron2}  with 1k warm-up iterations, 0.003 base learning rate, and learning rate decays at 60k and 100k iterations with a total of 120k training iterations.

\subsection{Matching Object Instances} \label{method:match}
\begin{figure}
    \centering
    \includegraphics[width=0.47\textwidth]{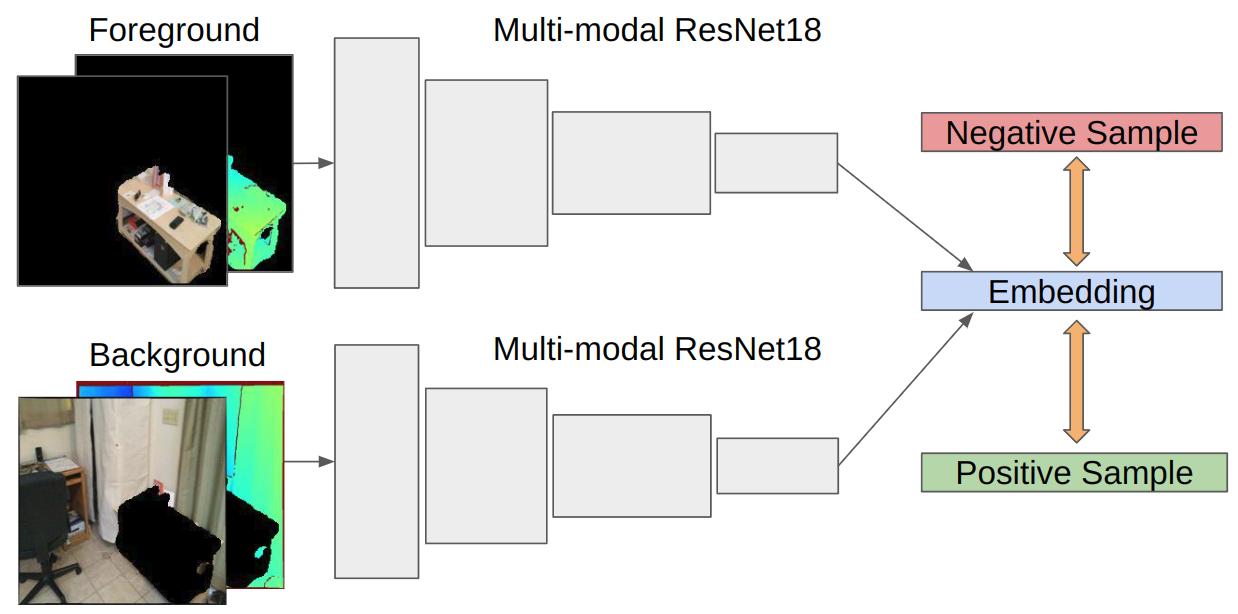}
    \vspace{-2mm}
    \caption{Our foreground/background metric-learning encoder for object matching, inspired by re-OBJ \cite{reid}. Using the detected and segmented objects from the model in Section \ref{method:noc}, we encode foreground and background regions of objects, using light-weight, multi-modal ResNet18 encoders on the RGB-D features.}
    \label{fig:match}
\end{figure}

In our global pose optimization formulation, the relation between frames is formed via a global identification of objects across frames. To enable such identification without any heuristic spatial assumptions, we train a simple metric learner that ranks object semantic similarities across frames.
Our object matching is shown in Figure~\ref{fig:match}.

We characterize objects for matching by their respective RGB-D features in the object instance mask, in addition to the global context in the 5-times upscaled object bounding box.
All inputs are resized to 224x224 and input to a lightweight ResNet18~\cite{resnet} backbone pre-trained on ImageNet~\cite{imagenet}. 

Similar to object detection, we employ two backbones for color and colored depth inputs. 
We omit normal input in object matching, as it empirically did not provide any benefit. For each input modality, we train two ResNet18 backbones for masked and inverse-masked crops, namely foreground (object) and background (context) encodings,
\begin{equation}
    e = \textrm{MLP}([\textrm{RN}^c(F_c), \textrm{RN}^c(B_c)] + [\textrm{RN}^d(F_d), \textrm{RN}^d(B_d)])
\end{equation}
where $\textrm{RN}$ are ResNet18s, $\textrm{MLP}$ is a fully connected network, $F, B$ are foreground and background crops for color (c) and depth (d), and $e$ is the object embedding vector.

Given an anchor object embedding $e_a$, a correctly matching object embedding $e_p$, and a negative example $e_n$, we train our metric learning using a triplet margin loss~\cite{triplet}:
\begin{equation}
L_{\textrm{tri}} = \textrm{max}(\textrm{d}(e_a, e_p) - \textrm{d}(e_a, e_n) + 1.0, 0),
\end{equation}
where $\textrm{d}$ is the $\ell_2$ distance.

We only consider triplets from the same category, as the object recognition pipeline provides classification. At inference time, we match the best instances using the Hungarian algorithm and apply a threshold $d(e_i, e_j) < \alpha$ for the matching object pairs from the same class. This semantic matching can be scaled to multiple frames via, e.g., object tracking with re-identification, or in our case, a simple pose graph optimization over frame pairs.

\smallskip
\noindent \textbf{Implementation.} We implement the identification network using PyTorch~\cite{pytorch} and train it on ScanNet data~\cite{scannet}. We train the network for 100k iterations with a batch size of 8 using a momentum optimizer with a learning rate 1e-4 and momentum 0.9.

\subsection{Energy Optimization} \label{method:opt}
We realize the joint energy minimization in Eq.~\ref{optim_high_level} using keypoint and NOC constraints. 

Using the predicted NOC constraints with back-projected depths, we can re-write the $\textrm{E}_o$ in Eq.~\ref{optim_high_level} as:
\begin{equation}\label{eq:eo}
    \textrm{E}_o(T, \bar{T}, P^{\textrm{\textrm{depth}}}, P^{\textrm{noc}}) = \sum_{o}\sum_{c}\sum_{k} || T_{c} P^{\textrm{depth}}_{o,c,k} - \bar{T}_{o} P^{\textrm{noc}}_{o,c,k} ||_2^2
\end{equation}
where $\bar{T}$, $T$ represent 9-DoF and 6-DoF camera and object transformations, respectively, and subscripts $o, c, k$ correspond to objects, cameras (frames), and points (pixels) within the frames, respectively. Here, object indices are determined by the object identification, and global object poses $\bar{T}$ indirectly constrain frames to each other without any explicit inter-frame feature matching.

In many cases, object constraints alone may not be sufficient to optimize the camera pose (e.g., frames may not share matching objects together). 
However, our object-based optimization is fully complementary to classical feature matching, and we thus formulate our objective in combination with feature-matching constraints $\textrm{E}_c$:
\begin{equation}
    \textrm{E}_c(T, P^{\textrm{depth}}) = \sum_{i}\sum_{j} \sum_{m, n} ||T_{i} P^{\textrm{depth}}_{i,m} - T_{j} P^{\textrm{depth}}_{j,n}||_2^2
\end{equation}

Our method is agnostic to the feature matcher, whether classical or learning-based. 
In this work, we experiment with two different keypoint matching systems to realize $\textrm{E}_c$, namely SuperGlue \cite{superpoint, superglue} and Geometric Transformer \cite{geotrans}, both offering state-of-the-art indoor feature matching.

With both object and feature-matching constraints, we realize the desired joint energy formulation as
\begin{equation}
    T^{*}, \bar{T}^{*} = \textrm{argmin}_{T, \bar{T}}(w_c\textrm{E}_c + w_o\textrm{E}_o).
\end{equation}
where $w_c, w_o$ weight feature-matching, and object energies.

Since non-linear least squares problems can be sensitive to outliers, we additionally employ outlier removal. 
Similar to BundleFusion~\cite{bundlefusion}, we apply Kabsch filtering to both intra-frame and keypoint constraints, using the matching depth-NOC and depth-depth correspondences, respectively. 
That is, we iteratively solve an orthogonal Procrustes problem and only keep the correspondences that have lower optimization errors. We use a liberal 20cm threshold to handle wide-baseline frames. Objects and inter-frame matches are rejected if the number of NOCs is $< 15$ and the number of keypoints is $ < 5$, respectively.

To solve this least-squares problem, we use a Gauss-Newton optimizer. To handle outliers during this global optimization, we remove $>15cm$ error residuals during optimization. As global optimization can produce coarse results, we further apply an ICP refinement. 

To handle SLAM-style sequence registration, we use a state-of-the-art global pose graph optimization~\cite{global_reg, open3d} over pairwise frame registration, using a single hierarchy level for simplicity. 
However, our method could be scaled to multi-frame optimization via object tracking and subdivided into multiple hierarchy levels to handle very large-scale scenes.

\section{Experiments}
\subsection{Pairwise Registration}\label{subsec:pairwise}
\noindent\textbf{Evaluation Dataset.} We evaluate pairwise camera pose estimation results on 1569 challenging and diverse frame pairs from the validation set of ScanNet~\cite{scannet}.
Unlike previous works~\cite{superglue, predator}, our evaluation includes a significant number of low-overlap pairs, with over 15\% of pairs having $\leq10\%$ overlap.

\medskip\noindent\textbf{Optimization.} To establish object correspondences across frame pairs in Eq.~\ref{eq:eo}, we use the top-1 matching object with embedding distance threshold 0.05. 
 In the absence of both keypoint matches and object matches, we consider object matches with distance $<0.15$.
 Only objects of the same class label and predicted maximum scale ratio $<1.5$ can match.
 We determine the best matching object based on the number of NOC-depth constraints after the filtering described in Section~\ref{method:opt}.  We combine it with keypoint constraints and refine the results with ICP.
 
\medskip\noindent\textbf{Classical Baselines.} We compare our method against various hand-crafted feature-matching and registration baselines. These baselines include BundleFusion keypoint optimization with SIFT descriptors and Kabsch filtering~\cite{sift, bundlefusion}, (SIFT + BF), and two geometric global registration baselines, Fast Global Registration (Fast GR)~\cite{fast_global_reg} and Global Registration (GR)~\cite{global_reg}, based on approximate  and exact  RANSAC over FPFH~\cite{fpfh} features, respectively.  As a post-processing step, all methods are refined using ICP. We use Open3D~\cite{open3d} implementations for global registration baselines as well as ICP post-processing of all methods. Our custom implementation of SIFT + BF uses the SIFT implementation of OpenCV~\cite{opencv_library}.
 
\medskip\noindent\textbf{Learned Baselines.} We also compare with pose estimators using learned models for feature or object matching: Geometric Transformer (GeoTrans)~\cite{geotrans}, which performs dense geometric feature matching, and SG + BF which leverages the powerful learned SuperGlue~\cite{superglue} and SuperPoint~\cite{superpoint} feature matching approaches in combination with BundleFusion~\cite{bundlefusion} Kabsch filtering and 3D Gauss-Newton optimization.
 We also use our network predictions to create a 3D object tracking baseline, \emph{Object Track}, mimicking tracking-based object-SLAM systems. Object Track obtains local object poses and object matches from Sections~\ref{method:noc} and~\ref{method:match}, but instead uses relative local object poses to get the camera pose instead of a global energy optimization.
 All methods are refined using ICP as a post-processing step.
 
\medskip\noindent\textbf{Evaluation Metrics.} We use the Pose Recall metric for pose evaluation, following  previous RGB-D pairwise registration works~\cite{deep_global_reg, predator}. To comprehensively capture performance, we employ several thresholds for absolute translation (in cm) and rotation (in $^\circ$). Since absolute translation is more difficult than rotation, its thresholds are twice the angle thresholds, following~\cite{deep_global_reg}.

\begin{figure}
    \centering
    \includegraphics[width=0.45\textwidth]{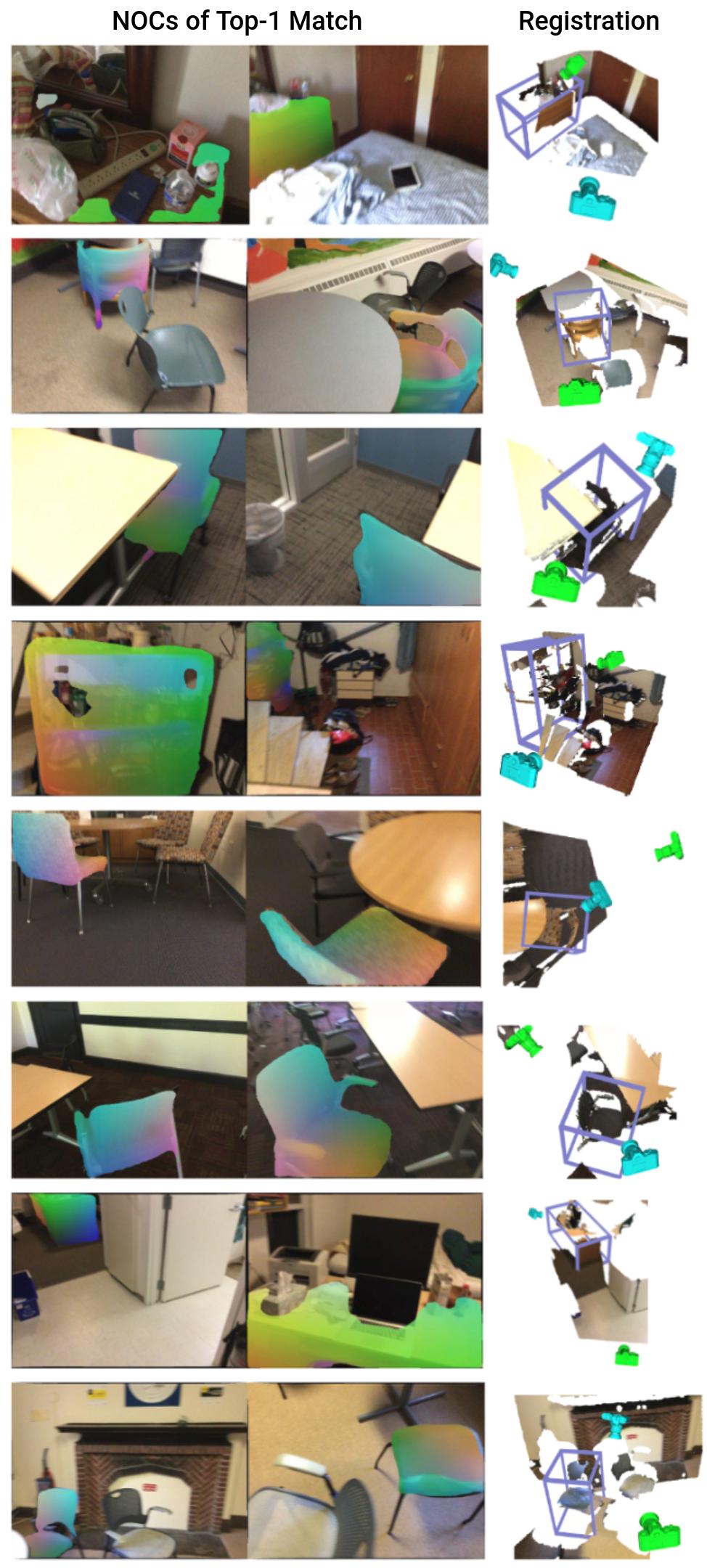}
    \vspace{-2mm}
    \caption{Low-overlap registration on ScanNet~\cite{scannet}, where traditional feature matching fails. Predicted NOC correspondences are visualized, along with object box and camera poses of the left and right images in green and blue, respectively.}
    \label{fig:s1_extreme}
\end{figure}

\begin{table}[htp]
\centering
\resizebox{0.45\textwidth}{!}{
\begin{tabular}{c|c c c}
     Method & \multicolumn{3}{c}{Pose Recall by Threshold} \\
     & 5$^\circ$, 10cm & 10$^\circ$, 20cm & 15$^\circ$, 30cm \\
     \hline
     SIFT + BF~\cite{sift, bundlefusion} & 16.83 & 18.42 & 20.40 \\
     Fast GR~\cite{fpfh, fast_global_reg} & 39.07 & 42.96 & 44.68 \\
     GR~\cite{global_reg} & 43.08 & 47.67 & 50.03 \\
     \hline
     Object Track & 37.41 & 43.21 & 45.19 \\
     GeoTrans~\cite{geotrans} & 67.11 & 74.57 & 76.93 \\
     SG + BF~\cite{superpoint, superglue, bundlefusion} & 71.13 & 80.11 & 81.71 \\
     \hline
     Ours (w/o keypoints) & 56.41 & 63.67 & 66.99 \\
     Ours (w/ GeoTrans) & 68.32 & 76.80 & 79.35 \\
     Ours (w/ SG + BF) & \textbf{74.25} & \textbf{84.58} & \textbf{87.06} \\
\end{tabular}
}
\vspace{-2mm}
\caption{Pose Recall results on ScanNet~\cite{scannet} validation images. Combined with state-of-the-art feature matching, our method outperforms various classical and learning-based baselines. 
Our approach complements both Geometric Transformer~\cite{geotrans} (Ours (w/ GeoTrans)) and SuperGlue~\cite{superglue} (Ours (w/ SG + BF)) feature matches, notably improving pose recall.
}
\label{S1_Results}
\bigskip
\centering
\resizebox{0.40\textwidth}{!}{
\begin{tabular}{
    c|c c c
}
     Method & \multicolumn{3}{c}{Recall by Overlap \%} \\
     & $\leq10$ & $(10, 30)$ & $\geq30$ \\
     \hline
     SIFT + BF~\cite{sift, bundlefusion} & 1.30  & 2.20 & 33.59\\
     Fast GR~\cite{fpfh, fast_global_reg} & 0.00 & 13.97 & 69.04 \\
     GR~\cite{fpfh, global_reg} & 0.65 & 20.36 & 74.62 \\
     \hline
     Object Track & 8.44  & 31.54 & 58.86 \\
     GeoTrans~\cite{geotrans} & 22.08 & 64.27  & 93.11  \\
     SG + BF~\cite{superpoint, superglue, bundlefusion} & 24.03 & 73.45 & 95.95  \\
     \hline
     Ours (w/o keypoints) & 32.47  & 53.29 & 80.31 \\
     Ours (w/ GeoTrans) & 35.06 & 65.47 & 94.42  \\
     Ours (w/ SG + BF) & \textbf{45.45} & \textbf{81.44} & \textbf{97.16}\\
\end{tabular}
}
 \vspace{-2mm}
\caption{Recall at 15$^\circ$, 30cm by overlap percentage. Our method significantly outperforms strong baselines in challenging low-overlap frame pairs, almost doubling recall for overlap $\leq$10\%. }
\label{tab:overlap}
\end{table}
 
\begin{figure}
    \centering
    \includegraphics[width=.42\textwidth]{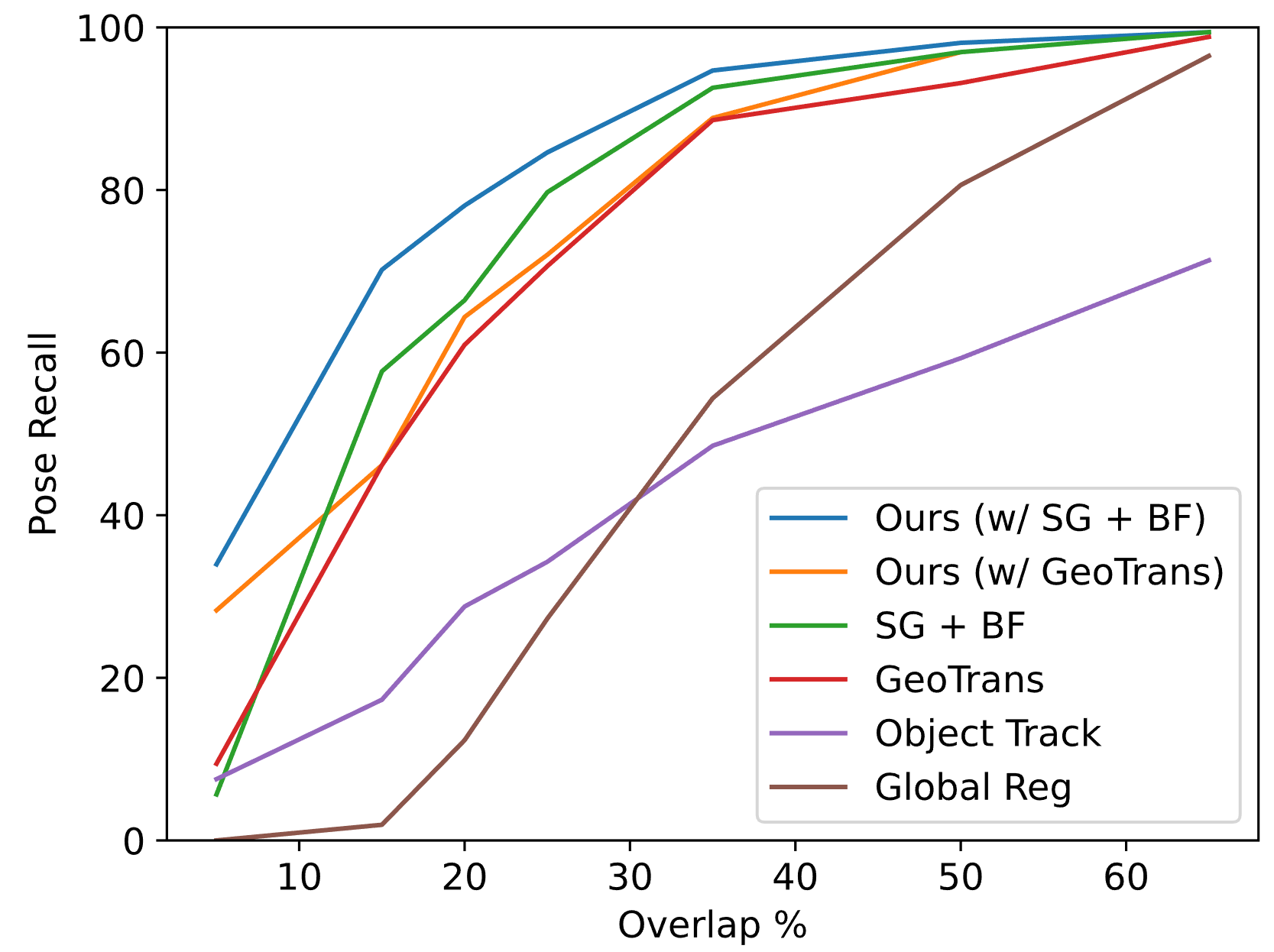}
    \vspace{-3mm}
    \caption{Change of pose performance by overlap percentage on ScanNet~\cite{scannet} validation pairs. Our method is significantly more robust against decreasing overlap compared to various classical and learned baselines. }
    \label{fig:overlap}
\end{figure}

\medskip\noindent\textbf{Quantitative Results.} In Table~\ref{S1_Results} and Table~\ref{tab:overlap}, we evaluate on ScanNet~\cite{scannet} validation frame pairs, measuring recall in various thresholds and overlap levels. Combined with state-of-the-art feature matching in a joint optimization, our method outperforms both classical and learning-based methods. Furthermore, the gap increases with lower overlap, as shown in Table~\ref{tab:overlap} and Figure~\ref{fig:overlap}, since our method can leverage object-based correspondences to estimate alignment in the absence of keypoint matches.
 We also show that our global optimization (Ours (w/o keypoints)) significantly outperforms naive object-based tracking (Object Track), demonstrating the efficacy of our global optimization formulation.
 Our approach is designed to complement keypoint matching, notably improving the state of the art in combination with state-of-the-art geometric and RGB-D feature matching.

\medskip\noindent\textbf{Analysis of Performance by Overlap.} In Table~\ref{tab:overlap} and Figure~\ref{fig:overlap}, we show 30cm, 15$^\circ$ recall performance at different geometric overlap rates, using the radius-based geometric overlap percentage measurement from~\cite{predator} with 1cm threshold. While all methods' performances decrease with overlap, our method retains significantly more robustness. 
Classical keypoint matching (SIFT + BF, Fast GR, SIFT) suffer strongly in low-overlap scenarios, which are challenging for hand-crafted descriptors.
The learned approaches of GeoTrans and SG+BF maintain some performance with decreasing overlap but still suffer strongly.
Our incorporation of object-based reasoning effectively complements keypoint matching while significantly improving robustness to low overlap, obtaining a performance improvement from 24.03\% to 45.45\% for overlap $\leq10\%$.

\medskip\noindent\textbf{Qualitative Results.} Figure~\ref{fig:s1_extreme} shows examples of low-overlap registration on ScanNet~\cite{scannet} frame pairs. 
Due to the minimal overlap, traditional keypoint matching cannot find sufficient correspondences, while our object grounding enables accurate camera pose estimation through indirect object constraints.

\subsection{Registration of RGB-D SLAM Sequences}\label{subsec:sequence}
\begin{table*}[ht]
\centering
\resizebox{0.99\textwidth}{!}{
    \begin{tabular}{c|c c c c c| c c c c c c c c c c c c}
    Method & \multicolumn{5}{c}{TUM Scene @ 1Hz} & \multicolumn{12}{c}{ScanNet Scene @ 1.5Hz} \\
    & fr1/desk & fr1/xyz & fr2/xyz & fr2/desk & fr3/office 
    & 0011 & 0081 & 0169 & 0207 & 0423 & 0430 & 0461 & 0494 & 0718 & 0760 & 0773 & 0805 \\
    \hline

    \hline
    SIFT + BF~\cite{sift, bundlefusion} & 30.60 & 3.49  & 2.73 & 157.02 & 38.01
    & 222.49  & 108.03 & 126.57 & 104.64 & 160.69 & 248.00 & 20.90 & 33.98 & 81.48 & 38.03 & 22.47 & 86.80 \\
    Redwood~\cite{fpfh, global_reg, open3d} & 3.10 & 1.97 & 1.88 & 140.17 & 175.64
    & 164.01 & 50.96 & 47.24 & 47.17
    & 88.43 & 156.76 & 10.47 & 9.77 & 133.53
    & 6.55 & 43.19 & 253.39\\
    SG + BF~\cite{superpoint, superglue, bundlefusion} & \textbf{3.01} & \textbf{1.92} & \textbf{1.81} & 5.30 & 5.14 &
    13.88 & 6.77 & 8.09 & 5.62 & 7.67
    & 35.97 & 3.31 & 7.16 & 13.42 & 6.37 & 12.43 & 12.98\\
    \hline
    Ours (w/ SG + BF) & \textbf{3.01} & \textbf{1.92} & \textbf{1.81} & \textbf{5.26} & \textbf{4.92} &
    \textbf{12.53} & \textbf{5.73} & \textbf{6.85} & \textbf{5.32} & \textbf{4.75} & \textbf{18.23} & \textbf{2.67} & \textbf{6.85} & \textbf{6.70} & \textbf{5.88} & \textbf{7.00} & \textbf{10.86}\\
    
    \end{tabular}
    }
    \vspace{-2mm}
    \caption{ATE RMSE values (cm) on room-scale TUM-RGBD~\cite{tum_rgbd} and ScanNet~\cite{scannet} test scenes. We evaluate at challenging 1 FPS and 1.5 FPS by sampling every 30th and 20th frame from TUM-RGBD and ScanNet, respectively. \OURS{} outperforms strong baselines leveraging classical and learning-based SLAM systems. }
    \label{tab:slam}
\end{table*}

We further demonstrate the effectiveness of our method in an RGB-D SLAM setting, registering sequences of frames in room-scale scenes.
In particular, we evaluate challenging, low-frame-rate settings that reflect common non-expert capture such as fast camera motion. 

For optimization, we use a global pose graph optimization over pairwise registration similar to the multi-way global registration of~\cite{global_reg, deep_global_reg, open3d}, as it offers an off-the-shelf, simple, and robust solution. We refer to our supplementary material for further optimization details.

\medskip\noindent\textbf{Datasets and Evaluation Metrics.} We evaluate our method on TUM-RGBD~\cite{tum_rgbd} and ScanNet~\cite{scannet} scenes. 
For ScanNet, we use a set of 12 scenes from the validation and test sets, having a wide range of sizes, environments, and camera trajectories.
We sample every 30th frame for TUM-RGBD (1Hz) and every 20th frame in ScanNet (1.5Hz). 
All methods use the same hyperparameters for each dataset, respectively.
To evaluate registration quality, we use the standard root mean squared trajectory error used for TUM-RGBD evaluation~\cite{tum_rgbd}.

\medskip\noindent\textbf{Baselines.} We implement various popular feature descriptors from cutting-edge SLAM systems in global pose graph optimization setups and thoroughly tune the hyper-parameters of each method to support low-frame-rate sequences. We compare against Redwood~\cite{global_reg}, a global multi-way pose graph that uses FPFH~\cite{fpfh} features. We also deploy robust RGB-D SIFT features used in BundleFusion~\cite{sift, bundlefusion} in a global pose graph optimization, creating the SIFT + BF baseline. Finally, we use the state-of-the-art SuperGlue matches of SuperPoint features with BundleFusion's sparse 3D optimization, SG + BF, as described in Section~\ref{subsec:pairwise}. 

\medskip\noindent\textbf{Quantitative Results.} In Table~\ref{tab:slam}, we compare our method to state-of-the-art traditional and learned SLAM approaches.
Across both TUM-RGBD and ScanNet validation and test scenes~\cite{tum_rgbd, scannet}, our object-grounded approach enables improved camera pose estimation. 
In particular, for the larger-scale ScanNet scenes, \OURS{} achieves notable reductions in trajectory error, of  38\%, 49\%, 50\%, and 44\% in scenes 0423, 0430, 0718, and 0773.

\begin{figure}
    \centering
    \includegraphics[width=0.47\textwidth]{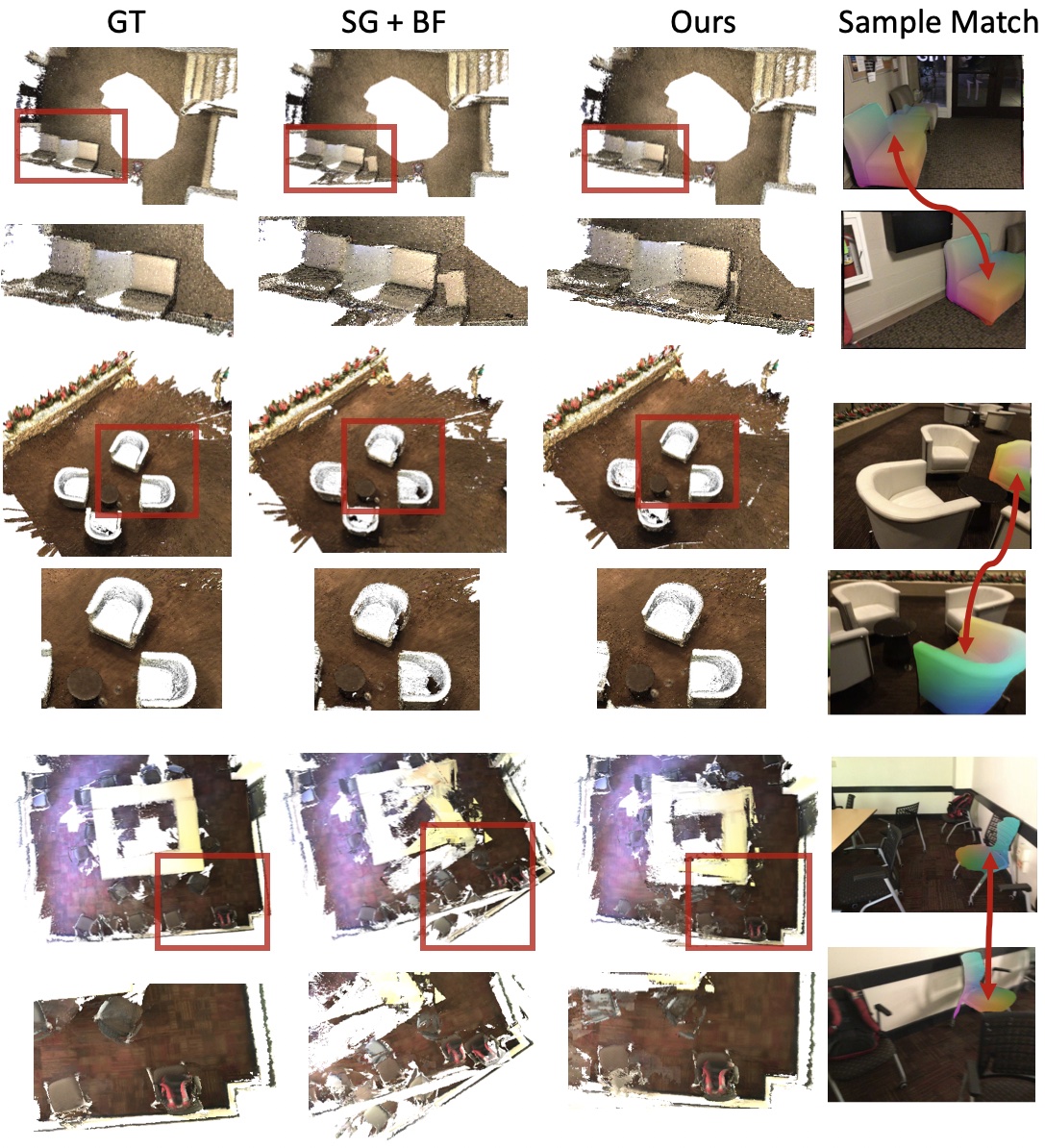}
    \vspace{-2mm}
    \caption{Qualitative comparison for registration of ScanNet scenes~\cite{scannet}, in comparison with only feature matching SG + BF\cite{superpoint, superglue, bundlefusion}. 
    All results are visualized by volumetric fusion of the respective RGB-D frames with the estimated poses.
    Our optimized poses  produce more accurate, clean, and consistent reconstruction, due to incorporating the object-based information in our global optimization.
    Example loop closures with matching object correspondences are shown on the right. 
    }
    \label{fig:slam_res}
\end{figure}

\medskip\noindent\textbf{Qualitative Results.} In Figure~\ref{fig:slam_res}, we show ScanNet~\cite{scannet} scene reconstructions using estimated camera poses.
RGB-D frames are fused using a scalable TSDF volumetric fusion~\cite{curless1996volumetric,open3d}.
Our approach complements state-of-the-art feature matching~\cite{superglue, superpoint, bundlefusion}, enabling more accurate and consistent reconstructions using our indirect correspondences, in comparison to feature matching alone.
We refer to the supplemental for additional qualitative sequence registration results.

\medskip\noindent\textbf{Limitations and Future Work.}
\OURS{} shows the capability of incorporating object semantics into global pose optimization; however, various limitations remain.
In particular, our approach leverages higher-level semantic correspondence given by objects for registration; however, in scenarios where object matches do not exist in frame views, we cannot leverage this constraint and instead use only feature matches.
Further leveraging background information regarding shared structures seen from different views would provide additional semantic correspondence information to further complement registration.
Additionally, while estimating canonical correspondences with objects can provide significant view information, object alignments can be somewhat coarse, while reasoning at a finer-grained level of object parts would provide more precise reasoning.
Finally, we believe future work can reduce our method's system complexity  by adopting a joint multi-frame SLAM optimization with fewer hyperparameters as well as a unified end-to-end architecture for learning object correspondences and identification together.

\section{Conclusion}
We have presented \OURS{} which introduces a new paradigm for incorporating object semantics into camera pose registration.
We leverage indirect, canonical object correspondences established with normalized object coordinates for camera pose optimization aspects in RGB-D registration and SLAM. 
To obtain these correspondences, we propose two multi-modal neural networks for object recognition with normalized object coordinate prediction and object identification.
\OURS{} operates in tandem with state-of-the-art feature matching in a joint Gauss-Newton optimization, with its object grounding  enabling registration to handle frame pairs with very low to no shared visibility. 
As a result, \OURS{} significantly improves the state-of-the-art feature matching in both pairwise and sequence registration of RGB-D frames, particularly in the challenging low-overlap regime.
Overall, we hope our method opens up new possibilities in the context of leveraging semantic information for camera pose estimation.

\medskip\noindent\textbf{Acknowledgements.}
This work was supported by the ERC Starting Grant Scan2CAD (804724), the Bavarian State Ministry of Science and the Arts coordinated by the Bavarian Research Institute for Digital Transformation (bidt), the German Research Foundation (DFG) Grant ``Making Machine Learning on Static and Dynamic 3D Data Practical'', and the German Research Foundation (DFG) Research Unit ``Learning and Simulation in Visual Computing.''



{\small
\bibliographystyle{ieee_fullname}
\bibliography{main}
}

\appendix

\section{Supplementary Overview}
In this supplementary material, we first show additional registration results in Section~\ref{s:qual} and additional ablation studies in Section~\ref{s:ablation}. We then describe additional method and baseline details in Section~\ref{s:method}.

\section{Additional Results}\label{s:qual}

\begin{figure}[h!]
    \centering
    \includegraphics[width=0.47\textwidth]{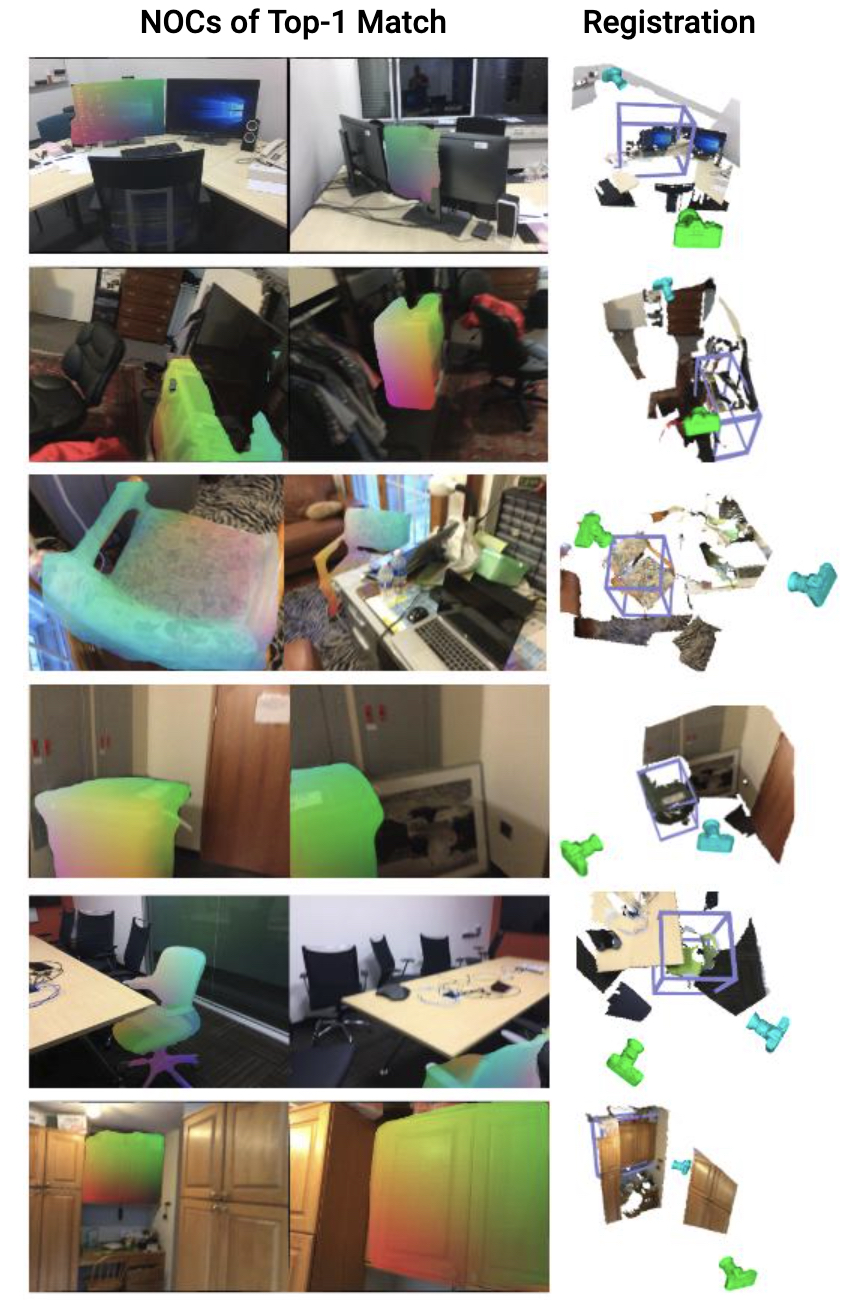}
    \vspace{-2mm}
    \caption{Additional low-overlap registration on ScanNet~\cite{scannet}, where traditional feature matching fails. Predicted NOC correspondences are visualized, along with object box and camera poses of the left and right images in green and blue, respectively.}
    \label{fig:supp_extreme}
\end{figure}

\noindent\textbf{Additional Pair Results.} In Figure~\ref{fig:supp_extreme}, we show additional low-overlap view registration samples from ScanNet~\cite{scannet} validation and test images.

\begin{figure}
    \centering
    \includegraphics[width=.47\textwidth]{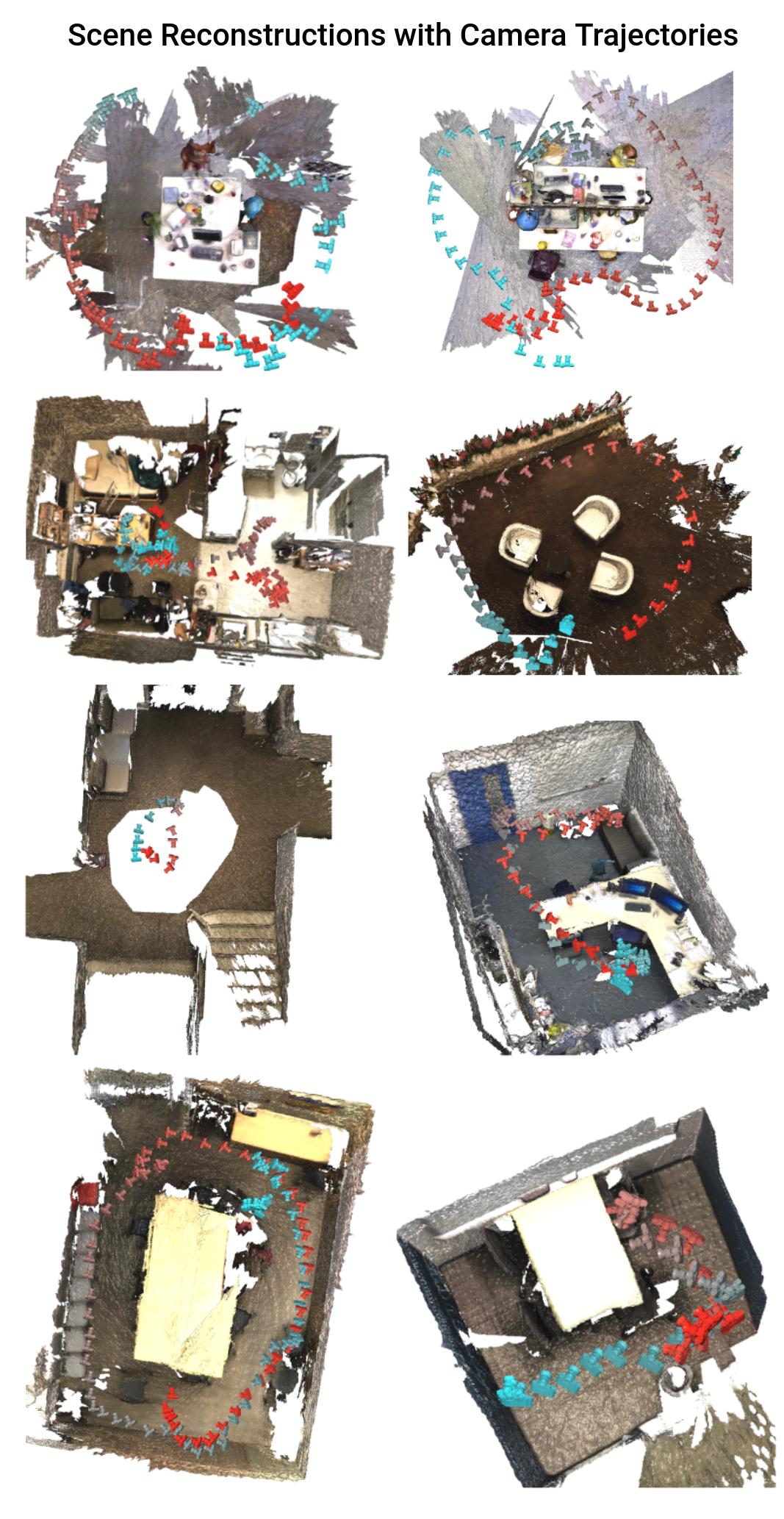}
    \vspace{-4mm}
    \caption{Example scene reconstructions from TUM RGB-D~\cite{tum_rgbd} and ScanNet~\cite{scannet} with optimized camera trajectories. Interpolation of camera color from blue to red represents the temporal order of cameras. Geometric reconstructions are obtained using scalable TSDF volume integration~\cite{curless1996volumetric, open3d}.}
    \label{fig:slam}
\end{figure}

\medskip\noindent\textbf{SLAM Reconstructions.} In Figure~\ref{fig:slam}, we show various scene reconstructions from TUM RGB-D~\cite{tum_rgbd} and ScanNet~\cite{scannet} using our camera pose estimates with a scalable volume integration~\cite{curless1996volumetric, open3d}. We show that our method obtains consistent and high-quality reconstructions in challenging low-frame-rate scenarios. We use 1 fps for TUM RGB-D (top row) and 1.5 fps for ScanNet (last 3 rows).

\medskip\noindent\textbf{Runtime.} On an RTX 3090, our method takes 0.63s for pairwise registration (except I/O, 0.05s SuperGlue) and can be made orders of magnitude faster by  JIT and custom kernels for network inference and GN optimization. 

\medskip\noindent\textbf{Pairwise Evaluation on SuperGlue Test Split.} We also evaluate our model on ScanNet test pairs introduced in SuperGlue~\cite{superglue}. While SuperGlue evaluation pairs are sampled with a bias towards higher-overlap pairs (only 19\% of pairs have $<30\%$ overlap, less than 4\% of pairs for $<10\%$ overlap), our method does not suffer but even improves pose recall on this data, as shown in Table~\ref{tab:superglue}.

\begin{table}[h!]
\centering
\resizebox{0.40\textwidth}{!}{
\begin{tabular}{
    c|c c
}
     Method & \multicolumn{2}{c}{Recall by Overlap \%} \\
     & $<30\%$ & $\geq30\%$ \\
     \hline
     SG + BF~\cite{superpoint, superglue, bundlefusion} & 77.39 & 98.19 \\
     Ours (w/ SG + BF) & \textbf{80.21} & \textbf{98.52} \\
\end{tabular}
}
\vspace{-2mm}
\caption{Registration recall by overlap in the SuperGlue~\cite{superglue} ScanNet~\cite{scannet} test pairs @ 15$^\circ$, 30cm.
}
\label{tab:superglue}
\end{table}

\begin{figure*}
\begin{center}
    \centering
    \includegraphics[width=.99\textwidth]{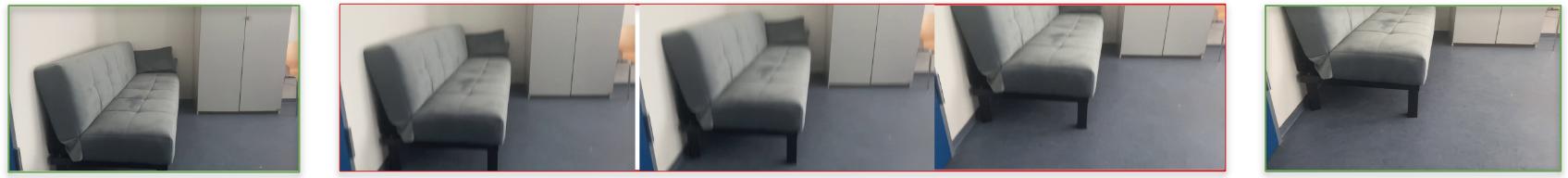}
    \vspace{-2mm}
    \caption{A self-captured sequence of frames with a smartphone. We show that two non-blurry frames on the left and right are apart by $>$ 20 frames, where motion blur is measured using the variance-of-Laplacian thresholding. We show three sample frames in between that are affected by the motion blur. Hence, low FPS registration would have real-world use in scenarios where frames not affected by motion blur must be considered for registration. }
    \label{fig:blur}
\end{center}

\end{figure*}

\medskip\noindent\textbf{Motion Blur in Non-Expert Capture Video.} To demonstrate the practical applicability of low FPS registration, in Fig.~\ref{fig:blur}, we show a sample non-expert captured real video sequence with motion blur where two sharp frames are more than 20 frames apart. 

\medskip\noindent\textbf{Low FPS Discussion.} Lower FPS registration can enable various applications, allowing for more compute budget for other tasks, or enabling the selection of low motion blur frames for reconstruction during fast capture. 
On the other hand, our approach remains beneficial in higher FPS scenarios, for instance improving ATE RMSE from 5.41 of SG+BF to 5.22 (ours) at 10fps on shorter ScanNet scene718\_00, and 21.67 to 16.37 at 3fps on longer ScanNet scene430\_00. To handle longer sequences and higher frame rates, we need to incorporate additional hierarchy levels, which we do not employ in this work for a simpler yet fair comparison between different feature-matching and object constraints.

\section{Ablation Studies}\label{s:ablation}
\noindent\textbf{Effect of Multi-modal Inputs on Object Recognition.} To measure the effect of multi-modal (color, depth, normals) learning in object recognition, we use a set of per-frame object alignment accuracy metrics over the whole ScanNet25k~\cite{scannet} validation set. We measure Scan2CAD~\cite{scan2cad} alignment accuracy over object poses as well as standard 2D recognition metrics, restricting the number of possible objects from a category per image instead of per scene. We show the results in Table~\ref{tab:multi-modal}. The local alignments are obtained using the input depths and predicted NOCs.
Our multi-modal approach shows notable benefits over color features only.

\begin{table}
\centering
\begin{tabular}{c|c c}
Input Modality & Class Avg. & Instance Avg. \\
\hline
Color & 37.33 & 41.31 \\
Color + Depth & 46.46 & 54.34 \\
Color + Depth + Normal & \textbf{48.92} & \textbf{55.09} \\
\end{tabular}
\vspace{-1mm}
\caption{Multi-modal inputs for object recognition. We evaluate Scan2CAD \cite{scan2cad} alignment accuracy over ScanNet25k validation images~\cite{scannet}. Adding jet-colored depth input~\cite{jet_multimodal} improves performance significantly. Normal input also offers a notable improvement, particularly in category average, helping infrequently-seen categories to generalize better.
}
\label{tab:multi-modal}
\end{table}

\medskip\noindent\textbf{Effect of Multi-modal Inputs on Pairwise Registration. } In Table~\ref{tab:pairmultimodal}, we measure the effect of multi-modal inputs in the final registration task, using our best method combined with SG+BF. Multi-modal inputs especially benefit in the low-overlap scenario, achieving a registration recall improvement from 37.01\% in the color-only case to 45\% when using depth and normal inputs.

\begin{table}
\centering
\resizebox{0.45\textwidth}{!}{
\begin{tabular}{c|c c c}
    Method & $\leq 10$ & (10, 30) & $\geq 30$ \\
    \hline
    GeoTrans & 22.08 & 64.27 & 93.11  \\
    SG + BF & 24.03 & 73.45 & 95.95 \\
    \hline
    Ours (Color only) & 37.01 & 77.84 & 96.50  \\
    Ours (Color + Depth only) & 42.21 & 80.64 & \textbf{97.16} \\
    Ours (final) & \textbf{45.45} & \textbf{81.44} & \textbf{97.16} \\
\end{tabular}
}
\vspace{-1mm}
\caption{Pose recall @ 30cm, 15$^\circ$ by overlap \% range. Ours refers to Ours (w/ SG + BF), with color, depth, and normal inputs. }
\label{tab:pairmultimodal}
\end{table}

\begin{figure}
    \centering
    \includegraphics[width=.47\textwidth]{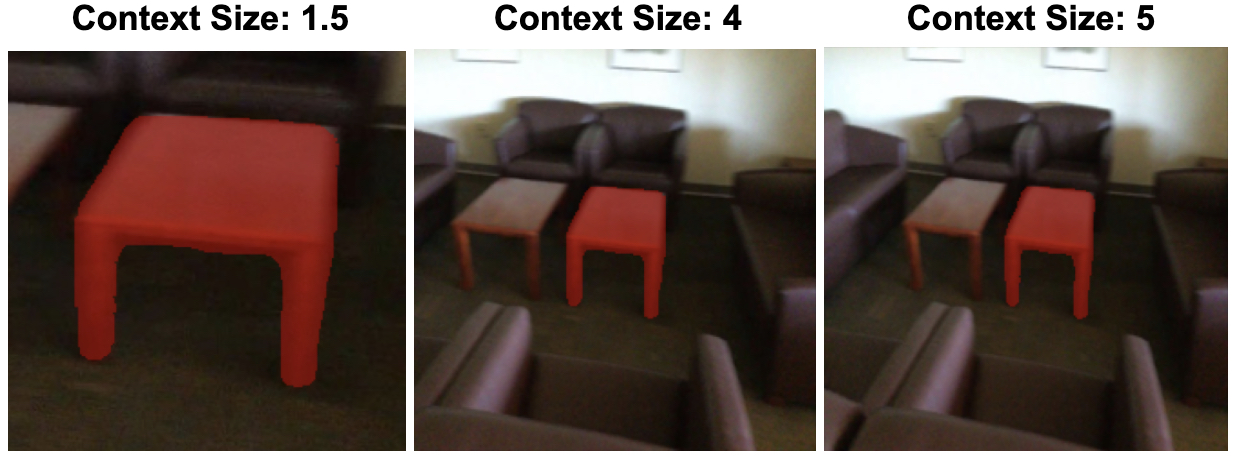}
    \vspace{-2mm}
    \caption{Crops of object regions in different context sizes, where object mask is magnified in red. A larger context size captures more details regarding the whole image while smaller context sizes capture more object details.}
    \label{fig:ctx}
\end{figure}

\medskip\noindent\textbf{Effect of Object Identification Context Size.} To assess the effect of context size (i.e., the scaling factor of the detected bounding box) on object matching, we evaluate the effect of context size top-1, top-2, and top-3 correctness of best-matching objects between ScanNet~\cite{scannet} validation pairs that have at least 1 shared object. top-2 and top-3 refer to up to 2 and 3, since all pairs may not have that many objects.
We consider three different context sizes: 4, 5, and 6. These context sizes $c$ are used to scale the detected object boxes $B$ as $cB$ when used to crop image regions for object identification.
We show the effect of context size in Figure~\ref{fig:ctx} and  Table~\ref{tab:context}, and find that $c=5$ produces the most robust performance.

\begin{table}
\centering
\begin{tabular}{|c|c c c|}
Context Size & top-1 & top-2 & top-3 \\
\hline
4 & 92.86 & 84.30 & \textbf{81.27} \\
5 & \textbf{93.41} & \textbf{85.40} & 80.72 \\
6 & 92.03 & 84.57 & 79.89 \\
\end{tabular}
\vspace{-1mm}
\caption{We evaluate top-1, top-2, and top-3 object matching accuracy using different context sizes. Context size is used to scale the detected object box for ROI-cropping for object identification.
We show a context size of 5 is a sweet spot for a robust top-1 and top-2 object matching, while a smaller context size is better for top-3 matching. 
}
\label{tab:context}
\end{table}

\medskip\noindent\textbf{Effect of Background Context in Object Identification.} Without encoding the background and training our identification model that only uses foreground object crops without bounding box scaling, top-1 object identification accuracy drops from 93.41\% to 91.21\%.

\medskip\noindent\textbf{Top-1 vs. Top-2 Filtering.} In Table~\ref{tab:top_n}, we evaluate the registration recall at different overlap levels for top-1 and top-2 matching object selection. To be selected, objects must be from the same category with mean embedding distance  $<0.05$. We do not report top-3 matching, since it produces identical results to top-2 due to our other outlier removal strategies. In low-overlap cases, top-1 matching offers additional robustness, while with high overlap the two strategies perform identically.
This is expected since in high-overlap frames, identical objects and their contexts look more similar to each other. On the other hand, in the low-overlap regime, there is a higher chance of having ambiguities since different objects may look similar due to viewpoints, and usually, there is a smaller number of matching objects compared to high-overlap cases. Therefore, having only the best-matching object increases the robustness by reducing the chance of error.

\begin{table}
\centering
\begin{tabular}{
    c|c c c
}
      & \multicolumn{3}{c}{Recall by Overlap \%} \\
     & $\leq10$ & $(10, 30)$ & $\geq30$ \\
     \hline
     Top-1 & \textbf{52.38} & \textbf{81.05} & \textbf{99.42} \\
     Top-2 & 42.86 & 77.89 & \textbf{99.42}  \\
\end{tabular}
 \vspace{-2mm}
\caption{Recall at 15$^\circ$, 30cm by overlap percentage. We compare top-1 and top-2 selections of objects in our best method variants, Ours (w/ SG + BF). We show top-1 matching is more robust, especially in the low-overlap cases.}
\label{tab:top_n}
\end{table}

\medskip\noindent\textbf{Effect of Symmetry Filtering in Pairwise Registration.} In Table~\ref{tab:sym}, we show the effect of filtering out the symmetric objects in our method, with and without keypoint constraints. While symmetry filtering helps in both cases, the effect is greater when keypoint constraints are not used, since keypoint constraints help to resolve potential ambiguities caused by the rotational symmetries.

\begin{table}
\centering
\resizebox{0.40\textwidth}{!}{
\begin{tabular}{c|c}
     Method & Pose Recall \\
     \hline
     Ours (w/o keypoint, w/o sym flt.) & 68.28 \\
     Ours (w/o keypoint) & \textbf{70.87} \\
     \hline\hline
     Ours (w/ SG + BF, w/o sym flt.) & 86.73 \\
     Ours (w/ SG + BF) & \textbf{87.38} \\
\end{tabular}}
\vspace{-1mm}
\caption{Pose recall @ 15$^\circ$, 30cm results with and without filtering out rotationally symmetric objects, where the filtering is enabled by rotational symmetry classification trained alongside NOC prediction. We show that symmetry filtering helps more in the object-only case (w/o keypoints), while still maintaining some improvement when our method is combined with keypoints. This shows that the keypoint constraints help to resolve symmetry ambiguities in object constraints. 
}
\label{tab:sym}
\end{table}

\section{Method and Baseline Details}\label{s:method}

\subsection{Data Preparation}
We use ScanNet~\cite{scannet} RGB-D frames along with CAD model annotations from Scan2CAD~\cite{scan2cad} to provide supervision for object NOCs.
To train the Mask R-CNN~\cite{maskrcnn}-based NOC prediction network, we use ScanNet400k, a subset of the 2.5m ScanNet RGB-D frames defined by ROCA \cite{scannet, roca}. 
To supervise object class categories and their 9-DoF poses, we use the Scan2CAD CAD alignment labels. Different from ROCA, we match the alignment labels to ScanNet's own instance labels, thereby obtaining NOCs and object labels via an inverse projection of RGB-D depth measurements instead of renderings of CAD models. 

To train the object identification network, we sample triplets of objects; each triplet is sampled from the same scene. To ensure wide baseline coverage, we take the positive samples that are at least 100 frames apart. We use the predictions matched with the ground-truth labels to obtain object crops, matched using the Hungarian algorithm over predicted boxes.

\subsection{Architecture}
\noindent\textbf{Additional Mask-RCNN Backbone Details.} We use the weights and configuration of Mask-RCNN-R50-FPN-3x \cite{maskrcnn, resnet, fpn, imagenet, coco} model from Detectron2 \cite{detectron2} as initialization. Our method predicts 32x32 masks instead of 28x28, to increase the resolution of our object correspondences; therefore, our method pools 16x16 feature grids for region proposals. Due to its fully convolutional nature, the default mask head pre-trained on COCO \cite{coco} can still be used for initialization. We use a batch size of 4 images with 128 region proposals each for training and fine-tune each layer of the backbone, except the first 2 layers.

\medskip\noindent\textbf{NOC Prediction Head.} We use a fully-convolutional network that predicts NOCs for every object. We use the same $16\times 16\times 256$ pooled features as in the mask prediction. The feature map is first processed by four $3\times 3$ convolutions with channel sizes of 256. The resulting feature map is then upsampled to $32\times 32\times 256$ using a single $3\times 3$ convolution that maps the channel size to 1024, followed by a pixel shuffle operator \cite{superres, pytorch}. The upsampled feature map is further processed using a single $3\times 3$ convolutions and two $1\times 1$ convolutions all with hidden sizes 256 modeling a shared MLP. The output is obtained via a final $1\times 1$ convolution that projects the feature map to the desired output channel size of 3. We use ReLU activations for each layer except the output, and a padding of 1 for all $3\times 3$ convolutions. Only NOC values of foreground pixels are considered for training and inference, using the values of ground-truth and predicted segmentation masks, respectively.

\medskip\noindent\textbf{Scale Regression Head.} We use a fully-connected network (MLP) for regressing 3D anisotropic object scales. We use the same $16\times 16$ feature map as in the NOC head. For efficiency, the network input is first downsampled to $8\times 8\times 256$, using a single $5\times 5$ convolution with a stride of 2, and then flattened. Then, we apply two fully-connected layers with a hidden size of 1024. For the 3D scale output, we use a per-category affine layer, similar to the scale prediction head of Vid2CAD \cite{vid2cad} and ROCA \cite{roca}. That is, the final layer regresses 3 scale values for each of the 9 categories and selects the correct category using the object classification. This enables learning category-specific weights and biases that model the different scale statistics of different categories, e.g., tables being much larger than trash bins or three-person sofas having a different aspect ratio than chairs. All hidden layers use ReLU activations.

\medskip\noindent\textbf{Symmetry Classification Head.} We use an MLP that is identical to the scale regression head except for the final layer dimension. We use a per-class affine output since the symmetry statistics of each category tend to differ, e.g., trash cans and tables are more often symmetric than chairs.

\medskip\noindent\textbf{Object Identification Network.} We use a metric learning approach for identifying objects across frames. Our model backbone is built from an ImageNet~\cite{imagenet}-pre-trained ResNet18~\cite{resnet} architecture provided by Torchvision~\cite{pytorch}. We use each layer except the final linear layer, and replicate the backbone for both background and foreground in each input modality (i.e., color and depth), without parameter sharing. We concatenate the foreground and background features, sum the concatenated color and depth features, and feed the result to an output network that applies a $2\times 2$ max-pooling, a $2\times 2$ convolution that doubles the feature channels, followed by a global max pooling, and an output linear layer that produces a 1024-dimensional embedding. We use ReLU activations for every hidden layer. We train the network using an initial learning rate of 1e-4, which is decreased by 10 when no improvement has occurred in the last two validation steps; validation is run every 5k iterations, evaluating top-2 matching accuracy between validation image pairs.

\medskip\noindent\textbf{Input Depth and Normal Preprocessing.} We use multi-modal networks~\cite{jet_multimodal} to process the depth inputs for both NOC prediction and object identification. For NOC prediction, we normalize the depth inputs assuming a maximum depth of 10 meters (sufficient for indoor rooms) and then color the depths using inverse jet coloring from Matplotlib~\cite{matplotlib} such that red represents near and blue represents far since orange/yellow tones are more common than blue tones in real indoor images. We also estimate normals from depths, using bilateral filtering followed by a nearest neighbor downsampling for depth, followed by a smooth normal estimation using $5\times 5$ Sobel filters. We observe that using normals with half the size of the depth map improves computational efficiency and model accuracy, likely due to their smoothness. Normals are directly colored to RGB by mapping the $[-1,1]$ range to $[0,255]$ range. For object identification, we also use inverse jet-colored depths. However, we normalize the depths using the maximum depth value observed in the image rather than using the absolute maximum depth. 

\subsection{Energy Optimization on RGB-D Sequences}
\noindent\textbf{Optimization for Temporal Sequence Registration.} 
Due to the temporal nature of sequence data, we apply adjusted thresholds from the pairwise registration scenario. 
In filtering consecutive frames, we use a 30cm instead of a 20cm threshold in BundleFusion Kabsch filtering~\cite{bundlefusion} threshold for keypoint matches to ensure consecutive frames can be registered, with other outlier removal thresholds remaining the same. For non-consecutive frames that may contain potential loop closures, we make the constraints stricter, using a 15cm threshold for BundleFusion Kabsch filtering. We also use a 0.04 threshold for object matching to ensure further robustness in loop closures. 

\medskip\noindent\textbf{Loop Closure Outlier Rejection.} We apply various filters to accept or reject loop closures, i.e., frame pair matches that are not consecutive. We only apply loop closure filters for ScanNet data~\cite{scannet}, since TUM RGB-D~\cite{tum_rgbd} scenes are relatively small.

When using loop closures with objects, we only accept loop closures where objects' optimized depths (z-dimension of the translation) are below 2.15m in at least one of the frames in a frame pair. We also ensure the $z$-dimension of the translation is positive in at least one of the matching frames. To apply such translation filtering, we transform the optimized global object poses to the local object poses using the optimized camera pose. We also filter out degenerate object optimizations by filtering out any object with an optimized scale dimension less than 0.05.

For loop closure without objects, we apply a global translation filtering by rejecting loop closures whose optimized relative camera translation is too large. That is, we allow a maximum of 60cm translation in nearby loop closures (within the consecutive 20 frames) and 1.5m for loop closures that are farther away. We exclude loop closure edges that do not adhere to these constraints from the global pose graph optimization.

\medskip\noindent\textbf{Pose Graph Optimization Details.} We use the default global Gauss-Newton pose graph optimizer from Open3~\cite{open3d, global_reg}. For ScanNet~\cite{scannet}, we use a maximum correspondence distance of 0.1, an edge prune threshold of 0.45, and use a 100\% preference rate for loop closures. For TUM RGB-D~\cite{tum_rgbd}, we use loop closure preference of 35\% with all other hyper-parameters the same.

\medskip\noindent\textbf{Pose Graph Restructuring.} We also make some adjustments to the standard pose graph structure of \cite{global_reg} to handle low-frame-rate scenarios more robustly. 
We make spatially distant consecutive edges uncertain, using a translation threshold of 40cm in TUM RGB-D~\cite{tum_rgbd} and 50cm in ScanNet~\cite{scannet}. We also make some non-consecutive edges certain if they have a translation of less than 4.5cm. This helps to overcome distant consecutive frames by assigning nearby loop closures as parents, helping significantly in performance.

\subsection{Baselines}
\noindent\textbf{GeoTransformer.} We evaluate the 3DMatch~\cite{3dmatch} pre-trained Geometric Transformer~\cite{geotrans}, since re-training or fine-tuning on our ScanNet~\cite{scannet} pairs did not empirically help in terms of further generalization. We use the standard LGR optimizer to obtain results since it worked slightly better than RANSAC in our experiments. When combining with our method, we integrate the weighted feature matches to our Gauss-Newton optimization following a BundleFusion-style~\cite{bundlefusion} outlier removal, similar to our handing of SuperGlue baseline~\cite{superpoint, superglue}.

\medskip\noindent\textbf{Redwood Optimization Details.} We use the best-performing pose-graph hyperparameters for the Redwood (Global Registration) \cite{global_reg} method to obtain maximum robustness in different datasets of SLAM sequences. First, we cancel the graph restructuring process since it does not provide any benefits. For ScanNet~\cite{scannet}, we use edge prune threshold 0.25 and max correspondence distance 1.0, with loop closure preference of 100\% for long sequences ($>$115 frames) and 30\% for other sequences. For TUM RGB-D, we change the edge prune threshold to 0.5 and the loop closure preference rate to 50\%. The baseline initially uses FPFH~\cite{fpfh} features with RANSAC, but it falls back to ICP for odometry edges with $<0.5$ convergence score.

\medskip\noindent\textbf{BundleFusion SIFT Optimization Details.} We use the best-performing hyperparameters for the SIFT + BF~\cite{sift, bundlefusion} approach for sequence registration. We use a Procrustes threshold of 50cm for ScanNet and 30cm for TUM RGB-D, respectively, for odometry cases, and 30cm for ScanNet and 20cm for TUM RGB-D for loop closures. We use max correspondence distance and edge pruning threshold of 0.5 for ScanNet and 0.5 and 3 for TUM-RGBD. We also use graph re-structuring in ScanNet, using a 4cm threshold to make loop closure edges certain. For both TUM-RGBD and ScanNet, we use a 1m threshold in nearby frames and 50cm and 75cm thresholds in far-away frames to make them uncertain. We refer to Open3D~\cite{open3d} registration pipelines for further detail.

\end{document}